\documentclass[letterpaper,10pt,conference]{ieeeconf}  

\IEEEoverridecommandlockouts                              

\usepackage{mymacros}
\usepackage{times} 
\usepackage{graphicx}

\usepackage[T1]{fontenc}
\usepackage{color}
\usepackage{pifont}

\usepackage{caption} 

\let\labelindent\relax
\usepackage{enumitem}

\title{\LARGE \bf 
    Mobile Manipulation Instruction Generation from Multiple Images with Automatic Metric Enhancement
}

\author{
    Kei Katsumata, Motonari Kambara, Daichi Yashima, Ryosuke Korekata, and Komei Sugiura
\thanks{
    \footnotesize The authors are with Keio University, 3-14-1 Hiyoshi, Kohoku, Yokohama, Kanagawa 223-8522, Japan.
    {\tt\footnotesize \{ke59ka77, motonari.k714, ydaichi1207, rkorekata, komei.sugiura\} @keio.jp}
}
}

\begin{document}

\makeatletter
\let\@oldmaketitle\@maketitle 
\renewcommand{\@maketitle}{\@oldmaketitle 
    \vspace{-4mm}
    \centering

}
\makeatother
\newcommand{\shiftDocumentDownWithMargins}[1]{%
  \addtolength{\topmargin}{#1}      
  \addtolength{\textheight}{-#1}   
  \addtolength{\footskip}{#1}      
  \addtolength{\textheight}{#1}    
}
\shiftDocumentDownWithMargins{3mm}
\maketitle
\vspace{-2mm}
\thispagestyle{empty}
\pagestyle{empty}

\begin{abstract}
We consider the problem of generating free-form mobile manipulation instructions based on a target object image and receptacle image.
Conventional image captioning models are not able to generate appropriate instructions because their architectures are typically optimized for single-image.
In this study, we propose a model that handles both the target object and receptacle to generate free-form instruction sentences for mobile manipulation tasks.
Moreover, we introduce a novel training method that effectively incorporates the scores from both learning-based and n-gram based automatic evaluation metrics as rewards.
This method enables the model to learn the co-occurrence relationships between words and appropriate paraphrases. 
Results demonstrate that our proposed method outperforms baseline methods including representative multimodal large language models on standard automatic evaluation metrics.
Moreover, physical experiments reveal that using our method to augment data on language instructions improves the performance of an existing multimodal language understanding model for mobile manipulation.
\end{abstract}

\vspace{-1mm}
\section{Introduction
}
\vspace{-1mm}

The advancement and deployment of service robots are essential in a variety of contexts such as elderly care facilities and daily support for disabilities.
In particular, the integration of service robots in elderly care facilities significantly reduces the burden on caregivers and addresses the growing demand driven by the rise in the elderly population.
Incorporating natural language understanding capabilities into robots would enhance their functionality and make them more user-friendly.
To enhance their understanding of natural language instructions, training models with datasets containing high-quality instructions are essential. 
Nonetheless, the construction of such datasets presents a significant challenge: the required mobile manipulation instructions are labor-intensive to produce.
This is because annotators are required to create sufficiently clear instructions based on multiple images.
In fact, instructions for mobile manipulation are often complex and lengthy\cite{qi2020reverie}.
Consequently, the ability to automatically generate high-quality instructions would be highly beneficial.

\begin{figure}[tb]
  \centering
  \begin{minipage}[b]{\linewidth}
    \centering
    \includegraphics[width=\linewidth]{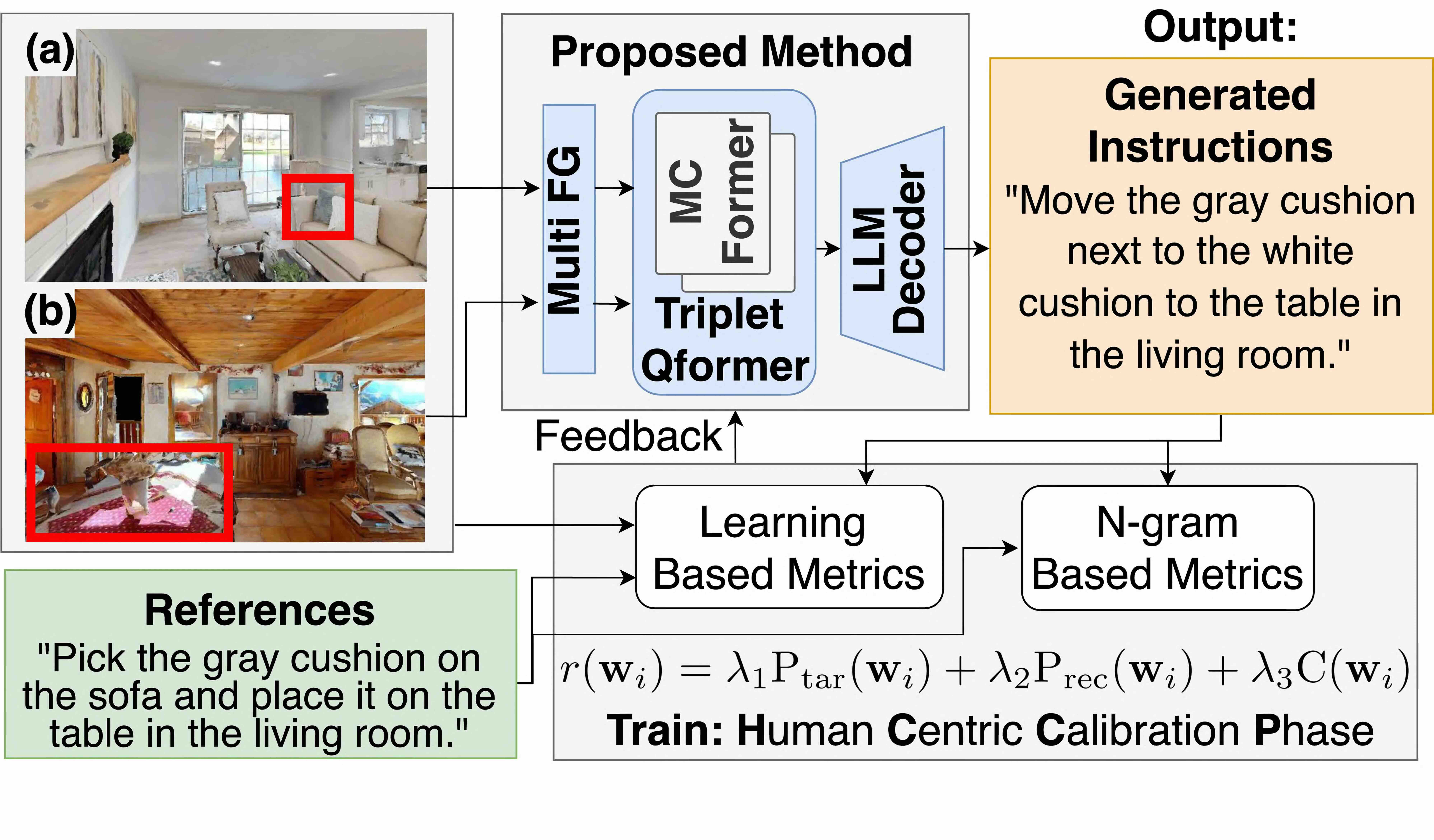}
  \end{minipage}
  \hfill
  \vspace{-10mm}
  \caption{Overview of our method. Our method generates a mobile manipulation instruction for a given target object image and receptacle image.}
  \label{fig:eye_catch}
  \vspace{-7mm}
\end{figure}
In this study, we focus on the task of generating free-form mobile manipulation instructions based on a target object image and receptacle image.
Fig. \ref{fig:eye_catch} shows a typical scene of this task.
In this example, the images in Figs. \ref{fig:eye_catch} (a) and (b) are the given target object image and the receptacle image, respectively. 
In addition, the target object is a gray cushion and the receptacle is a wooden table. We aim to generate a sentence such as ``Move the gray cushion next to the white cushion to the table in the living room.''
In this task, instructions need to be generated considering both the target object present in one image and the receptacle in the other image. 
Therefore, models are required to appropriately handle both images.
Most existing image captioning models (e.g.,\cite {nguyen2022grit, li2023blip2}) do not have architectures that can handle multiple images.
Hence, these methods are inappropriate for generating mobile manipulation instructions based on multiple images.

We propose a model that generates mobile manipulation instructions using a target object image and a receptacle image. 
An overview of our method is also shown in Fig.~\ref{fig:eye_catch}.
Our model identifies both the target object and receptacle, and it generates the corresponding instruction for the mobile manipulation task.
We introduce Triplet Qformer, which enables the alignment of multiple visual features with respect to text features.
Triplet Qformer embeds two types of visual features and one linguistic feature into the same space.
This enables the grounding of visual features in natural language for both target objects and receptacles.
Moreover, we introduce human centric calibration phase (HCCP), a training method that uses the human centric calibration training (HCCT) loss function in combination with learning-based and $n$-gram-based automatic evaluation metrics. 
Using HCCP, the model learns co-occurrence relationships between words and appropriate paraphrases of sentences.

The contributions of this study are as follows:
\begin{itemize}
    \item We introduce Triplet Qformer. This architecture enables multiple types of visual features to be aligned individually with text features. The text features are used as anchors to align the visual features with each other.
    \item We introduce HCCP, a training method that uses the HCCT loss function with learning-based and $n$-gram-based automatic evaluation metrics.
\end{itemize}
Our code is available at this URL.\footnote{https://github.com/keio-smilab24/MMIG}

\begin{figure*}[t]
    \centering
    \vspace{1.5mm}
    \includegraphics[width=\linewidth]{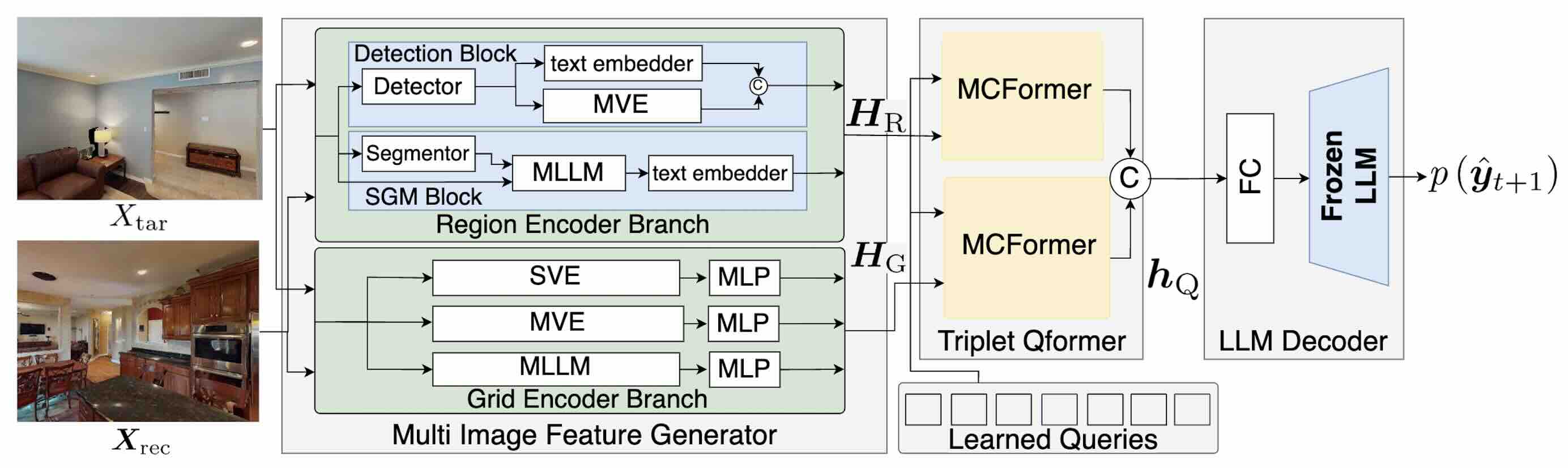}
    \vspace{-5.5mm}
    \caption{\small Overall architecture of the proposed model.
    \textcircled{c}, SVE, MVE, MLLM, and FC represent concatenation, single-modal visual encoder, multimodal visual encoder, multimodal large language model, and fully connected layer, respectively.}
    \label{fig:model}
    \vspace{-6.7mm}
\end{figure*}

\vspace{-1mm}
\section{
    Related Work
}
\vspace{-0.8mm}

\subsection{Multimodal Language Processing for Robotics}

Research related to vision and language in robotics is widely conducted, and several survey papers exist\cite{Firoozi2023FoundationMI, kawaharazuka2024real}.
For example, many benchmarks for mobile manipulation tasks have been proposed \cite{iocchi2015robocup, OkadaAR}. 
Most existing methods handle only template-based instructions or a limited number of fixed instruction types \cite{iocchi2015robocup, OkadaAR}. 
On the other hand, some free-form instruction-based mobile manipulation methods\cite{Driess2023palme, kaneda2024learning, korekata24arxiv} have also been proposed.

There are some standard datasets for the tasks with free-form natural language instruction(e.g., \cite{qi2020reverie}).
Matterport 3D (MP3D) \cite{Matterport3D} and Habitat-Matterport 3D (HM3D) \cite{ramakrishnan2021hm3d} are representative benchmarks in research focused on indoor environments.
These datasets are smaller in size compared to standard datasets used in other Vision \& Language tasks.
For example, the COCO dataset\cite{lin2014microsoft}, a standard for image captioning, contains approximately 1.7M sentences, and the average sentence length is 10.47 words \cite{hirsch2024clid}.
In contrast, the Remote Embodied Visual Referring Expression in Real Indoor Environments dataset\cite{qi2020reverie}, one of the representative datasets for mobile manipulation, contains only 21K sentences and the average sentence length is 18 words.
Thus, compared with datasets for other tasks, those with natural language instructions for mobile manipulation are not adequate in size to enable effective training.
Moreover, as can be inferred from the difference in average sentence length, instructions for mobile manipulation are often complex and lengthy. 
This complexity arises from the necessity to clarify tasks using referring expressions.
Therefore, constructing such datasets poses a challenge because of the labor-intensive and time-consuming nature of annotation.
\vspace{-1mm}

\subsection{Data Augmentation}
Data augmentation aims to create additional training data. 
Many methods have been proposed for the augmentation of captions\cite{xiao2023multimodal, yi2024augment}.
These approaches are often limited to word substitutions or paraphrasing. 
Nevertheless, text data can also be augmented using image captioning models.
Wang et al.\cite{wang2023scaling} generated large-scale data for VLN using photo-realistic environment. However, they found that the quality of captions in their work was limited.
\vspace{-2mm}
\subsection{Image Captioning}

Image captioning is the task of generating sentences based on a given image. 
Many image captioning methods have been proposed\cite{nguyen2022grit, li2023blip2}.
Recent developments in LLMs have enhanced their expressive capabilities\cite{li2023blip2, li2024evcap}.
BLIP-2 \cite{li2023blip2} is a representative captioning model employing an LLM decoder. 
It leverages a two-stage pretraining strategy to bridge the modality gap between vision and language.
However, these methods are not designed to handle two images simultaneously.

Many training methods have also been proposed for image captioning models.
In particular, methods that use automatic evaluation metrics as reward functions are widely employed \cite{nguyen2022grit, rennie2017self}.
Self-critical sequence training \cite{rennie2017self} aligns the training process with the evaluation metric, directly optimizing for CIDEr\cite{vedantam2015cider}.
Cho et al. \cite{cho-etal-2022-fine} also introduced a strategy that combines $n$-gram based metrics and learning-based metrics.
Though this approach has been successful in retrieval tasks, approaches that do not combine these metrics have remained more effective in image captioning tasks.
In this study, we introduce the novel training method HCCP, which effectively incorporates both learning-based and $n$-gram-based metrics into the loss function.

\vspace{-2mm}
\section{Problem Statements
}
\vspace{-1mm}

In this paper, we focus on the mobile manipulation instruction generation (MMIG) task. 
We define the OMIG task as follows: given two images, one of a target object and one of a receptacle, the model generates a free-form instruction sentence for fetching the target object and carrying it to the receptacle.
In this task, appropriate mobile manipulation instructions that include the target object and the receptacle based on the input images are to be generated.
Fig.\ref{fig:eye_catch} shows a typical scene from the OMIG task. In the images in Fig.\ref{fig:eye_catch} (a) and (b), the target object and the receptacle are a gray cushion and a table, respectively.
The goal is to generate an instruction sentence, such as ``Move the gray cushion next to the white cushion to the table in the living room.''
The terms used in this paper are defined as follows: 
A `target object' refers to an object needed in daily life, identified as the object to be grasped in the instruction, and a `receptacle' refers to a piece of furniture identified as the destination in the instruction where the target is to be placed.

The input for the task consists of an image of the target object (target object image) and the receptacle (receptacle image). Bounding boxes for specific objects are not provided. The output is a free-form mobile manipulation instruction, which includes both the target object and the receptacle.
In this study, we assume that the model is required to generate an instruction for a single pick-and-place task and path planning is not required for this task.
In addition, we assume that reference instructions are executable; in other words, if we can generate instructions similar to the reference, then the feasibility of execution is ensured.


\vspace{-0.5mm}
\section{Proposed Method
\label{method}
}
\vspace{-0.8mm}


We propose a model that generates mobile manipulation instructions by identifying the target object and receptacle in each image, respectively.
The proposed model consists of three main modules: Multi Image Feature Generator, Triplet Qformer, and LLM Decoder.
Fig. \ref{fig:model} shows the structure of our method. 
The input to our model is defined as $\bm{x} = \left\{\bm{X}_{\mathrm{tar}}, \bm{X}_\mathrm{rec} \right\}$.
Here, $\bm{X}_{\mathrm{tar}}$ and $\bm{X}_{\mathrm{rec}}$ denote a target object image and a receptacle image, respectively.
\vspace{-1mm}

\subsection{
    Multi Image Feature Generator
    \label{hfg}
}
Multi Image Feature Generator generates the region feature $\bm{H}_\mathrm{R}$, which includes local features of objects, and the grid feature $\bm{H}_\mathrm{G}$, which include the global features of the entire image, from $\bm{x}$.
Relying solely on either type of feature can lead to a lack of information regarding the details of objects or spatial relationships between objects. 
Therefore, using both types of features enables effective image representations.
This module consists of two branches: the region encoder branch and the grid encoder branch.

\subsubsection{\textbf{Region encoder branch}}

The region encoder branch computes the region features, which contain detailed information about objects.
The effectiveness of region features has been reported\cite{nguyen2022grit}. 
Most existing methods use single-modal region features based solely on image features.
Integrating linguistic information into region features can help to align text with image features more appropriately.
Therefore, our method uses multimodal region features that integrate image features obtained by the object detector with text features derived from the predicted labels.
Moreover, we leverage a multimodal LLM (MLLM) to obtain comprehensive textual descriptions of the objects in the image, including their spatial relationships and detailed characteristics.
The region encoder branch consists of two blocks: the detection block and the spatial grounded marks (SGM) block.

The region encoder branch is applied to each image $\bm{X}_\mathrm{tar}$ and $\bm{X}_\mathrm{rec}$ independently.  
The following explanation focuses on $\bm{X}_\mathrm{tar}$, because the same process is applied to $\bm{X}_\mathrm{rec}$. 
In this branch, we obtain two types of region features: the detection image feature $\bm{h}_\mathrm{{R,D}}$ in the detection block and SGM feature $\bm{h}_\mathrm{{R,S}}$ in the SGM block from $\bm{X}_\mathrm{tar}$.

First, in the Detection block, we obtain the sets of detected objects $\bm{D}_k$ with predicted labels using an object detector from $\bm{X}_\mathrm{tar}$.
Here, $k$ denotes the number of detected objects.
We provide the detector with a dictionary containing noun phrases extracted from the training data.
The dictionary is used to map the detector's outputs to the corresponding noun phrases, ensuring more appropriate association between image features of detected objects and text features from their labels.
Using a multimodal vision encoder (e.g., \cite{radford2021learning}), we extract image feature $\bm{v}_{\mathrm{{De}}, k}$ from $\bm{D}_k$.
Moreover, we extract text feature $\bm{s}_k$ from the predicted text labels of the detected objects.
Then, we obtain the feature $\bm{h}_\mathrm{{R,D}} = [\bm{v}_{\mathrm{De},k}; \bm{s}_k]$.

In the SGM block, we first apply marks to the images with segmentation masks using segmentation models and obtain descriptions using an MLLM from those images following a method similar to Set of Marks (SoM) \cite{yang2023setofmark}.
Then, we acquire the SGM feature $\bm{h}_\mathrm{{R,S}}$ from the description.
By executing the aforementioned procedure on $\bm{X}_{\mathrm{rec}}$, we obtain detection region feature $\bm{h}'_\mathrm{{R,D}}$ and the SGM feature $\bm{h}'_\mathrm{{R,S}}$.
Finally, we obtain the region feature $\bm{H}_\mathrm{{R}}$ as 
\begin{align}
\bm{H}_\mathrm{{R}} =  \left\{[\bm{h}_\mathrm{{R,D}}; \bm{h}'_\mathrm{{R,D}}], [\bm{h}_\mathrm{{R,S}}; \bm{h}'_\mathrm{{R,S}}]\right\}.
\end{align}
\vspace{-2.5mm}
\subsubsection{\textbf{Grid encoder branch}}
While the region features are helpful for capturing local object characteristics, they have a limited ability to provide a comprehensive representation of the image.
In many cases, region features do not include information about the space between objects, making it challenging to obtain positional relationships.
However, as previously mentioned, complex referring relationships, such as spatial relationships between objects, are crucial for identifying objects. 
To address these limitations, we introduce a grid encoder branch which computes grid features from the entire image.
The grid features are capable of acquiring global information, including the spatial relationships between objects.
The grid encoder branch generates three types of latent representations, and the grid features are obtained using three encoders: a single-modal visual encoder (e.g., DINOv2\cite{oquab2024dinov}), a pre-trained multimodal visual encoders (e.g., CLIP), and an MLLM (e.g., LLaVA\cite{liu2023llava}).
Each latent representation individually emphasizes different aspects: a single-modal visual encoder captures fine-grained visual details, a pre-trained multimodal visual encoder addresses structural and relational aspects, and the MLLM aligns these aspects with natural language for comprehensive understanding \cite{goko2024task}.
Combining these multimodal features is an effective way to generate mobile manipulation instructions using diverse expressions.
Grid features $\bm{h}_\mathrm{{G,D}}$, $\bm{h}_\mathrm{{G,C}}$, and $\bm{h}_\mathrm{{G,L}}$ are obtained from $\bm{X}_\mathrm{tar}$ as $\left(\bm{h}_\mathrm{{G,D}}, \bm{h}_\mathrm{{G,C}}, \bm{h}_\mathrm{{G,L}} \right) = \left( \mathrm{MLP}(\bm{v}_\mathrm{{D}}), \mathrm{MLP}(\bm{v}_\mathrm{{C}}), \mathrm{MLP}(\bm{v}_\mathrm{{L}}) \right)$,
 where $\bm{v}_\mathrm{D}$, $\bm{v}_\mathrm{C}$, $\bm{v}_\mathrm{L}$, and $\mathrm{MLP}\left( \cdot \right)$ denote the visual features extracted by a single-modal visual encoder, the visual features extracted by a multimodal encoder, the visual features extracted by an MLLM, and a linear transformation, respectively. 
Similarly, we obtain image features $\bm{h}'_\mathrm{{G,D}}$, $\bm{h}'_\mathrm{{G,C}}$, and $\bm{h}'_\mathrm{{G,L}}$ from $X_\mathrm{rec}$.
Finally, we obtain a grid feature 
\begin{align}
\bm{H}_\mathrm{{G}} =  \left\{ [\bm{h}_\mathrm{{G,D}}; \bm{h}'_\mathrm{{G,D}}], [\bm{h}_\mathrm{{G,C}}; \bm{h}'_\mathrm{{G,C}}], [\bm{h}_\mathrm{{G,L}}; \bm{h}'_\mathrm{{G,L}}] \right\}.
\end{align}

\vspace{-7mm}
\subsection{
    Triplet Qformer
    \label{tq}
}
\vspace{-0.8mm}
\begin{figure}[t]
    \centering
    \vspace{1mm}
    \includegraphics[width=0.95\linewidth]{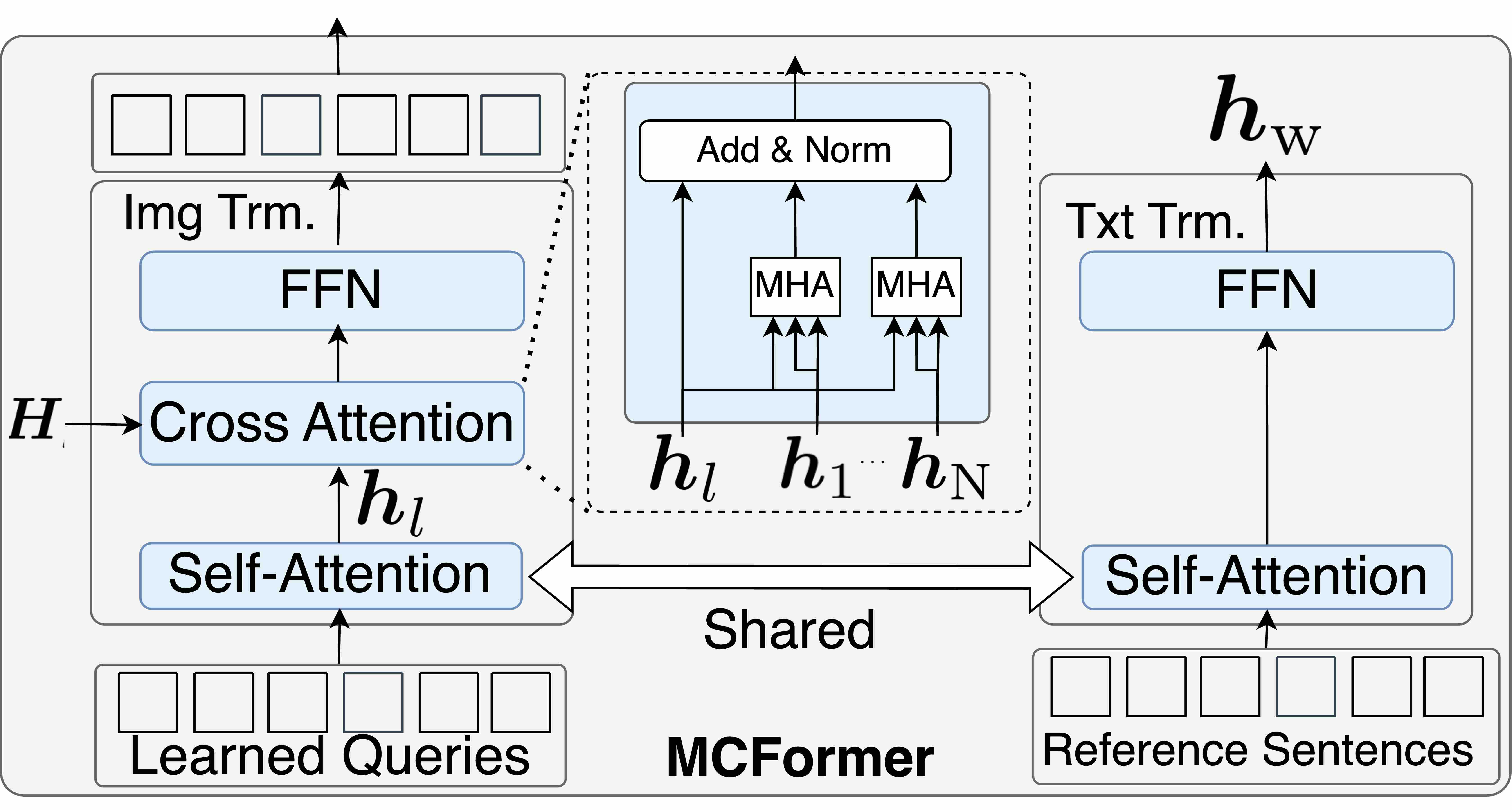}
    \vspace{-2.5mm}
    \caption{\small The details of MCFormer. Img Trm., Txt Trm., FFN and MHA represent image transformer, text transformer, feed-forward network and multi-head attention, respectively. Here, we define $\bm{H}$ as $\bm{H} = \left\{ \bm{h}_1, ..., \bm{h}_N \right\}.$ }
    \label{fig:mcformer}
    \vspace{-8mm}
\end{figure}
Next, we introduces Triplet Qformer, which aligns the two kinds of visual features with text features from the ground-truth description, using text features as the anchor.
This approach indirectly aligns the three types of features.
By aligning the entirety of the reference sentences with the object and receptacle features, the model can learn co-occurrence relationships between pairs of target objects and receptacles.
This module takes $\bm{H}_\mathrm{G}$, $\bm{H}_\mathrm{R}$, and learnable queries as inputs.
During pretraining, the reference sentence $\bm{y}$ is also used as input.
The output is multimodal feature $\bm{h}_\mathrm{Q}$.
Triplet Qformer consists of two Multi Image Cross Attention Transformers (MCFormers), 
the structure of which is shown in Fig. \ref{fig:mcformer}.
The first and second MCFormers use $\bm{H}_\mathrm{G}$ and $\bm{H}_\mathrm{R}$ as the image features, respectively.
For readability, the following describes a block that uses $\bm{H}_\mathrm{G}$ as the image feature.
MCFormer is an extension of Qformer\cite{li2023blip2}, which was designed to integrate multiple visual features and align these visual features with text features.

As shown in Fig. \ref{fig:mcformer}, the text and image features are aligned through the shared self-attention layers in the two transformers: the image transformer and the text transformer. 
For the image transformer, each layer includes a self-attention layer, a cross-attention layer, and a feed-forward network layer.
First, in the self-attention layer, we obtain $\bm{h}_l$ by applying self-attention to the learnable query embedding.
The parameters of the self-attention layer are shared with the self-attention layer in the text transformer.
As shown in Fig. \ref{fig:mcformer}, the cross-attention layer consists of $N$ multi-head attention (MHA) blocks.
Here, $N$ represents the number of elements in a set of image features. That is to say, $N = 3$ and $N = 2$ for $\bm{H}_\mathrm{G}$ and $\bm{H}_\mathrm{R}$, respectively.
Note that the parameters of each MHA block are not shared.
The output of each MHA block is denoted as $\bm{a}_n$, where \( n = 1, \ldots, N \).
We obtain an intermediate feature
\begin{align}
\bm{h}_\mathrm{{Q,G}} = \mathrm{FFN} \left( \mathrm{LN} \left( \sum_{n=1}^N \bm{a}_{n} + \bm{h}_l \right) \right),
\end{align}
where $\mathrm{LN}(\cdot)$ and $\mathrm{FFN}(\cdot)$ denote the layer normalization and feed-forward network, respectively.
By applying the same procedure to $\bm{H}_\mathrm{R}$, we also obtain the feature $\bm{h}_\mathrm{{Q,R}}$. 
Finally, we obtain intermediate feature $\bm{h}_\mathrm{Q} = [ \bm{h}_\mathrm{{Q,G}}; \bm{h}_\mathrm{{Q,R}}]$.

For the text transformer, each layer includes a self-attention layer and feed-forward network layer.
The output of the text transformer is $\bm{h}_\mathrm{w}$. This is used only during training, as discussed in Section \ref{trainingstages}.
\vspace{-2mm}

\subsection{
    LLM Decoder
    \label{llm}
}
\vspace{-1mm}
Existing image captioning models do not generalize well to out-of-domain images containing novel scenes or objects.
Therefore, we incorporate a pre-trained LLM as decoder. The LLM possesses highly expressive capabilities and an extensive vocabulary, which enhances its linguistic expressiveness.
LLM Decoder generates an instruction sentence based on $\bm{h}_\mathrm{Q}$.
In this module, we first use a fully connected layer to linearly project $\bm{h}_\mathrm{Q}$.
Then, we obtain the probability for the next token $p(\bm{y}_{t+1} | \bm{x}, \bm{y}_{1:t})$ from a frozen LLM. Here, $\bm{y}_{t+1}$, and $\bm{y}_{1:t}$ represent the predicted token at time $t+1$ and the predicted token sequence up to time $t$, respectively.


\vspace{-1mm}
\begin{table*}[hbt]
\centering

\setlength{\tabcolsep}{3.5pt}
\vspace{1mm}
\caption{Quantitative comparison between the proposed method and baseline methods on the test sets of the HM3D-FC dataset.Dense Caption refers to the method Dense Caption Baseline. Target and Receptacle correspond to the scores for the target object images and the receptacle images, respectively. The best score for each metric is in bold.}
\vspace{-1mm}
\normalsize
\begin{tabular}{>{\centering\arraybackslash}p{2.3cm}>{\centering\arraybackslash}p{1.42cm}>{\centering\arraybackslash}p{1.42cm}>{\centering\arraybackslash}p{1.42cm}>{\centering\arraybackslash}p{1.42cm}>{\centering\arraybackslash}p{1.42cm}>{\centering\arraybackslash}p{1.42cm}>{\centering\arraybackslash}p{1.42cm}>{\centering\arraybackslash}p{1.42cm}>{\centering\arraybackslash}p{1.42cm}} \hline
    \multirow{2}{*}{Method} & \multicolumn{2}{c}{Polos$\uparrow$} & \multicolumn{2}{c}{PAC-S$\uparrow$} & \multicolumn{2}{c}{RefPAC-S$\uparrow$} & \multirow{2}{*}{SPICE$\uparrow$} & \multirow{2}{*}{CIDEr$\uparrow$} & \multirow{2}{*}{BLEU4$\uparrow$} \\ 
     & \small{Target} & \small{Receptacle} & \small{Target} & \small{Receptacle} & \small{Target} & \small{Receptacle} &  & &  \\ \hline
     Dense Caption & \small{30.0 {$\scriptscriptstyle \pm 5.6$}} & \small{30.8 {$\scriptscriptstyle \pm 6.0$}} & \small{70.3 {$\scriptscriptstyle \pm 2.1$}} & \small{69.6 {$\scriptscriptstyle \pm 2.1$}} & \small{73.5 {$\scriptscriptstyle \pm 2.2$}} & \small{73.3 {$\scriptscriptstyle \pm 2.1$}} & \small{10.0 {$\scriptscriptstyle \pm 2.9$}} & \small{25.1 {$\scriptscriptstyle \pm 7.3$}} & \small{6.6 {$\scriptscriptstyle \pm 0.7$}}
    \\ \hline
    GRIT\cite{nguyen2022grit} & \small{41.2 {$\scriptscriptstyle \pm 1.8$}} & \small{39.0 {$\scriptscriptstyle \pm 1.8$}} & \small{70.8 {$\scriptscriptstyle \pm 1.4$}} & \small{67.6 {$\scriptscriptstyle \pm 1.9$}} & \small{75.5 {$\scriptscriptstyle \pm 1.1$}} & \small{73.7 {$\scriptscriptstyle \pm 1.6$}} & \small{19.8 {$\scriptscriptstyle \pm 0.5$}} & \small{59.2 {$\scriptscriptstyle \pm 2.8$}} & \small{10.5 {$\scriptscriptstyle \pm 0.3$}} \\ 
    BLIP-2\cite{li2023blip2} & \small{42.4 {$\scriptscriptstyle \pm 1.4$}} & \small{40.1 {$\scriptscriptstyle \pm 1.5$}} & \small{72.3 {$\scriptscriptstyle \pm 0.8$}} & \small{67.4 {$\scriptscriptstyle \pm 0.9$}} & \small{74.7 {$\scriptscriptstyle \pm 0.6$}} & \small{73.6 {$\scriptscriptstyle \pm 0.7$}} & \small{16.9 {$\scriptscriptstyle \pm 0.4$}} & \small{36.6 {$\scriptscriptstyle \pm 4.9$}} & \small{8.5 {$\scriptscriptstyle \pm 1.2$}} \\ 
    Gemini\cite{reid2024gemini} & \small{29.5 {$\scriptscriptstyle \pm 0.2$}} & \small{29.4 {$\scriptscriptstyle \pm 0.2$}} & \small{69.7 {$\scriptscriptstyle \pm 0.5$}} & \small{68.1 {$\scriptscriptstyle \pm 0.2$}} & \small{73.3 {$\scriptscriptstyle \pm 0.5$}} & \small{72.7 {$\scriptscriptstyle \pm 0.2$}} & \small{11.2 {$\scriptscriptstyle \pm 0.6$}} & \small{26.2 {$\scriptscriptstyle \pm 1.6$}} & \small{5.2 {$\scriptscriptstyle \pm 0.4$}} \\ 
    GPT-4o\cite{achiam2023gpt} & \small{34.5 {$\scriptscriptstyle \pm 0.2$}} & \small{35.5 {$\scriptscriptstyle \pm 0.06$}} & \small{72.1 {$\scriptscriptstyle \pm 1.4$}} & \small{70.7 {$\scriptscriptstyle \pm 0.5$}} & \small{75.4 {$\scriptscriptstyle \pm 0.8$}} & \small{75.0 {$\scriptscriptstyle \pm 0.7$}} & \small{15.2 {$\scriptscriptstyle \pm 0.5$}} & \small{33.4 {$\scriptscriptstyle \pm 1.5$}} & \small{7.5 {$\scriptscriptstyle \pm 0.2$}} \\
    Ours & \small{\textbf{50.9 {$\scriptscriptstyle \pm 0.8$}}} & \small{\textbf{50.7 {$\scriptscriptstyle \pm 0.7$}}} & \small{\textbf{74.3 {$\scriptscriptstyle \pm 0.6$}}} & \small{\textbf{72.1 {$\scriptscriptstyle \pm 0.7$}}} & \small{\textbf{79.8 {$\scriptscriptstyle \pm 0.4$}}} & \small{\textbf{78.5 {$\scriptscriptstyle \pm 0.5$}}} & \small{\textbf{22.7 {$\scriptscriptstyle \pm 0.4$}}} & \small{\textbf{64.8 {$\scriptscriptstyle \pm 4.0$}}} & \small{\textbf{11.5 {$\scriptscriptstyle \pm 1.1$}}} \\ \hline
\end{tabular}
    \vspace{-5.5mm}
\label{tab:result}
\end{table*}

\vspace{3mm}
\subsection{Training Stages
    \label{trainingstages}
}
\vspace{-1.5mm}

Our model is trained in three stages: Triplet Qformer pre-training phase (TQPP), probability distribution matching phase (PDMP), and human centric calibration phase (HCCP).
In TQPP, we pre-train Triplet Qformer using a loss function similar to that used in Qformer\cite{li2023blip2}.
Next, in PDMP, we train our entire model using the cross-entropy loss function.
Finally, in HCCP, we train the model selected in PDMP using the human centric calibration training (HCCT) loss function.
The HCCT loss function employs both $n$-gram and learning-based automatic evaluation metrics.
Existing methods\cite{nguyen2022grit, rennie2017self} have proposed loss functions that utilize $n$-gram-based metrics.
This enables them to learn the co-occurrence relationships between words present in the reference sentences.
However, several studies have reported that the performance of $n$-gram-based metrics is insufficient\cite{wada2024, sarto2023positive}.
In particular, it is challenging to appropriately evaluate paraphrases of reference sentences.
Some learning-based automatic evaluation metrics \cite{wada2024, sarto2023positive} are reported to have a higher correlation coefficient with human evaluations than those $n$-gram-based metrics.
The evaluation metrics based on the similarity between the generated sentences and images can appropriately evaluate paraphrases.
Therefore, by incorporating both learning-based and $n$-gram-based metrics into the loss function, the frozen LLM decoder's expressiveness should be leveraged more effectively for generating instruction sentences.
\vspace{-0.8mm}

We use different loss functions for each stage.
In TQPP, we employ the loss function $\mathcal{L}_\mathrm{TQPP}$.
We delineate the loss function in the training of the block that handles $\bm{h}_\mathrm{G}$.
 Here, $\mathcal{L}_\mathrm{TQPP}$ is defined as 
 \begin{align}
 \mathcal{L}_\mathrm{TQPP} = \mathcal{L}_\mathrm{itc} + \mathcal{L}_\mathrm{CE}(\bm{y}, \bm{h}_\mathrm{w}) +
 \mathcal{L}_\mathrm{BCE}(\bm{h}_\mathrm{Q,G}),
 \end{align}
where $\mathcal{L}_\mathrm{CE}(\cdot)$, $\bm{y}$, $\bm{h}_\mathrm{w}$, and  $\mathcal{L}_\mathrm{BCE}(\cdot)$ denote the cross-entropy function loss, a reference sentence, an output text feature of MCFormer, and the binary cross-entropy function loss, respectively.
We use the image-text contrastive loss $\mathcal{L}_\mathrm{itc}$ defined in \cite{li2023blip2}.
In PDMP, we employ the cross-entropy function.
Furthermore, in HCCP, we use the HCCT loss function $\mathcal{L}_\mathrm{HCCT}$.
$\mathcal{L}_{\mathrm{HCCT}}$ is defined as follows:
\begin{align}
\mathcal{L}_\mathrm{HCCT} = - \frac{1}{k} \sum_{i=1}^{k} (r(\mathbf{w}_i) - b) \log p(\mathbf{w}_i).
\end{align}
\noindent Here, $\mathbf{w}_i$, $r(\mathbf{w}_i)$, $b$, and $k$ denote the $i$-th generated sentence in the beam, reward function, reward baseline, and index of the sample in the batch, respectively.
Moreover, $r(\mathbf{w}_i)$ and $b$ are defined as follows:
\begin{align}
r(\mathbf{w}_i) &= \lambda_1 \mathrm{P}_{\mathrm{tar}}(\mathbf{w}_i) + \lambda_2 \mathrm{P}_{\mathrm{rec}}(\mathbf{w}_i) + \lambda_3 \mathrm{C}(\mathbf{w}_i), \nonumber \\
b &= \frac{1}{k} \sum_{i=1}^{k} r(\mathbf{w}_i),
\end{align}
where $\mathrm{P}_\mathrm{tar}(\cdot)$, $\mathrm{P}_\mathrm{rec}(\cdot)$, and $\mathrm{C}(\cdot)$ represent the Polos score for the target object image, the Polos score for the receptacle image, and the CIDEr\cite{vedantam2015cider} score, respectively.
Here, $\lambda_1$, $\lambda_2$, and $\lambda_3$ are hyperparameters.

\vspace{-1.5mm}
\vspace{-0.3mm}
\section{
    Experimental Settings
}
\vspace{-1mm}

In the experiments, we used the Habitat-Matterport 3D Dataset (HM3D) subset of the Learning-To-Rank in Real Indoor Environments for Fetch-and Carry (LTRRIE-FC) dataset (HM3D-FC \cite{korekata24arxiv}).
The dataset for the OMIG task should include target object images, receptacle images, and instruction sentences corresponding to each image pair.
Most standard datasets for the VLN task (e.g., \cite{qi2020reverie}) include information related to path planning in instructions. Such datasets were not appropriate for this study because we did not address path planning.
By contrast, the HM3D-FC dataset includes instruction sentences that are acceptable for our task.
Thus, we used the HM3D-FC dataset.

 In our experiments, the reward weights in the human centric calibration training loss function, we use 0.25, 0.25, and 0.5 for $\lambda_1$, $\lambda_2$, and $\lambda_3$, respectively.
 We trained our model on a GeForce RTX4090 with 24 GB of GPU memory.
It took approximately 16 hours in total to train our model.
The inference time was approximately 92 ms/sample.

In the region feature branch in Multi Image Feature Generator, we employed the following pre-trained models.
In the Detection block, we employed Detic\cite{zhou2022detecting}, CLIP, and RoBERTa\cite{liu2019roberta} as the detector, multimodal visual encoder for the detected object images, and text embedder for the labels of detected objects, respectively. Additionally, in the SGM block, we used SEEM\cite{Zou23seem}, GPT-4V\cite{achiam2023gpt}, and embedding-large-3\cite{text-embedding-3-large} as the segment model, MLLM, and text embedder, respectively.
In the grid feature branch, DINOv2, CLIP, LLaVA were used for single-modal visual encoder, multimodal encoder, and MLLM, respectively.
In LLM Decoder, we used OPT-2.7B\cite{zhang2022opt} as the frozen LLM.

\begin{figure*}[hbt]
    \centering
    \small
    \hspace{-2mm}
    \begin{minipage}[t]{0.32\textwidth}
        \centering
        \begin{tabular}{cc}
            \hspace{-5mm}
            \begin{tabular}{c}
                \includegraphics[width=0.5\linewidth]{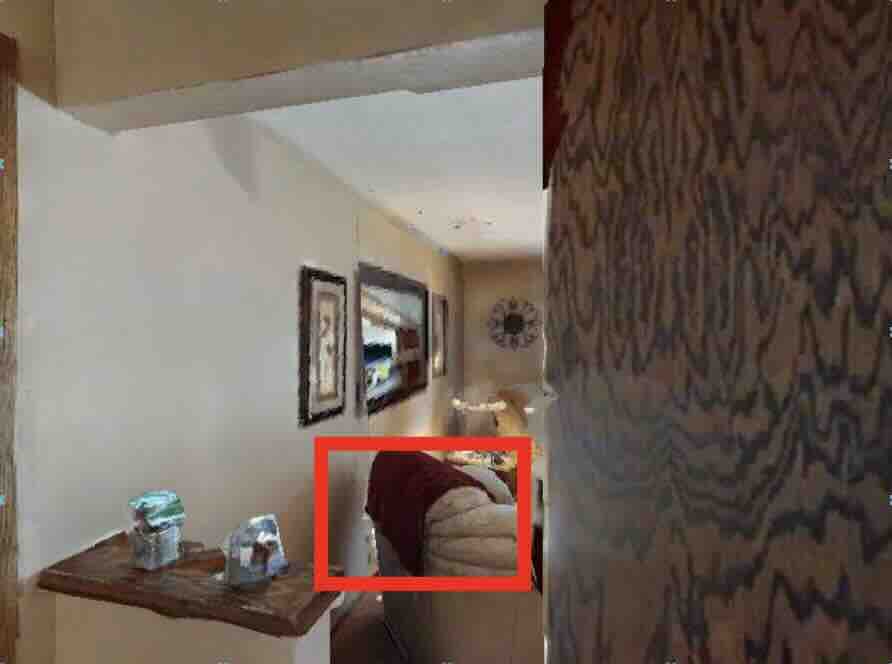} \\
                \textbf{(a)}
            \end{tabular} &
            \hspace{-8mm}
            \begin{tabular}{c}
                \includegraphics[width=0.5\linewidth]{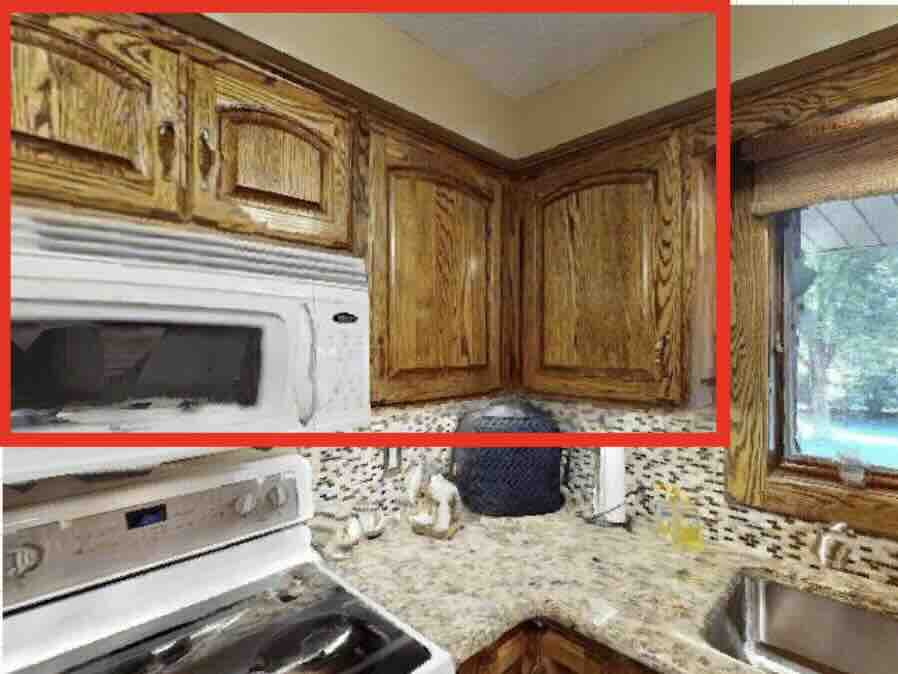} \\
                \textbf{(b)}
            \end{tabular}
        \end{tabular}
        \begin{tabular}{|p{0.95\linewidth}|}
            \hline
            \textbf{Ref:} ``Move the red object on the sofa to the cupboard at the corner.'' \\
            \\
            \hline
            \textbf{BLIP-2:} ``Move the white curtain on the left side of the window to the white shelf on the right side of the window.'' \\
            \hline
            \textbf{GPT-4o:} ``Take the two glass figurines from the wooden ledge and place them on the kitchen counter next to the sink.'' \\
            \hline
            \textbf{Ours:} ``Move the red object on the sofa to the shelf above the kitchen.'' \\
            \\
            \hline
        \end{tabular}
        \vspace{-6mm}
        \begin{tabular}{p{0.95\linewidth}}
        \vspace{-3mm}
        \begin{center}
        $\mathrm{(i)}$     
        \end{center}
        \end{tabular}
    \end{minipage}
    \hspace{1.5mm}
    \begin{minipage}[t]{0.32\textwidth}
        \centering
        \begin{tabular}{cc}
        \hspace{-6mm}
            \begin{tabular}{c}
                \includegraphics[width=0.5\linewidth]{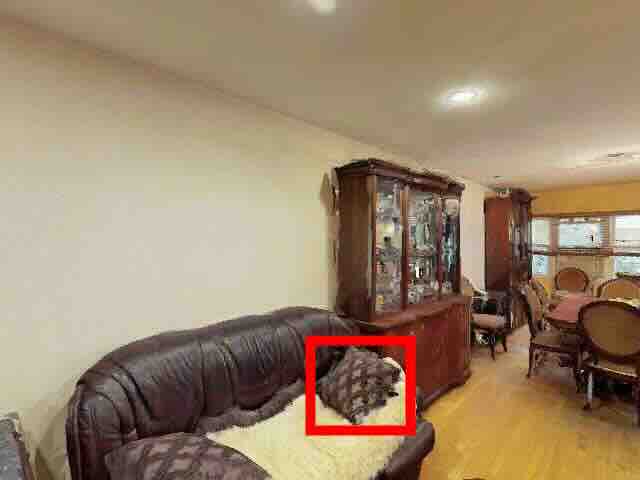} \\
                \textbf{(a)}
            \end{tabular} &
            \hspace{-8mm}
            \begin{tabular}{c}
                \includegraphics[width=0.5\linewidth]{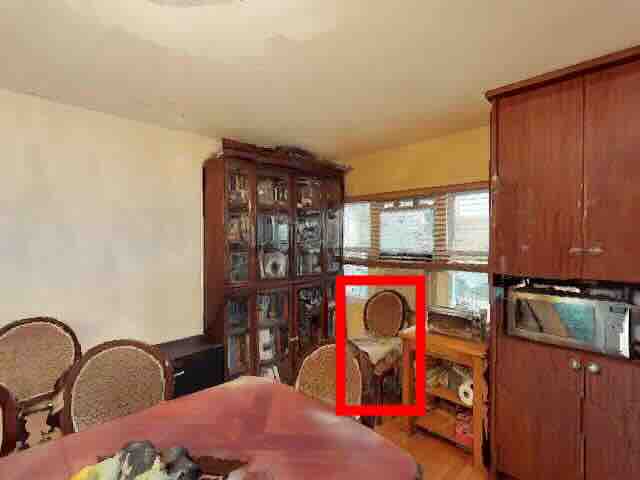} \\
                \textbf{(b)}
            \end{tabular}
        \end{tabular}
        \vspace{3mm}
        \begin{tabular}{|p{0.95\linewidth}|}
            \hline
            \textbf{Ref:} ``Pick up the cushion on the sofa and put it on the chair near the window.'' \\ 
            \\
            \hline
            \textbf{BLIP-2:} ``Could you move the cushion on the sofa and put it on the brown sofa?'' \\
            \\
            \hline
            \textbf{GPT-4o:} ``Take the brown cushion from the couch and place it on the counter next to the red bottle in the kitchen.'' \\
            \hline
            \textbf{Ours:} ``Could you move the brown cushion on the sofa to the wooden chair in the corner of the room?'' 
            \\
            \hline
        \end{tabular}
        \vspace{-2mm}
        \vspace{-1mm}
        \begin{tabular}{p{0.95\linewidth}}
        \vspace{-3mm}
        \begin{center}
        $\mathrm{(ii)}$     
        \end{center}
        \end{tabular}
    \end{minipage}
    \hspace{1.5mm}
    \begin{minipage}[t]{0.32\textwidth}
        \centering
        \begin{tabular}{cc}
        \hspace{-6mm}
            \begin{tabular}{c}
                \includegraphics[width=0.5\linewidth]{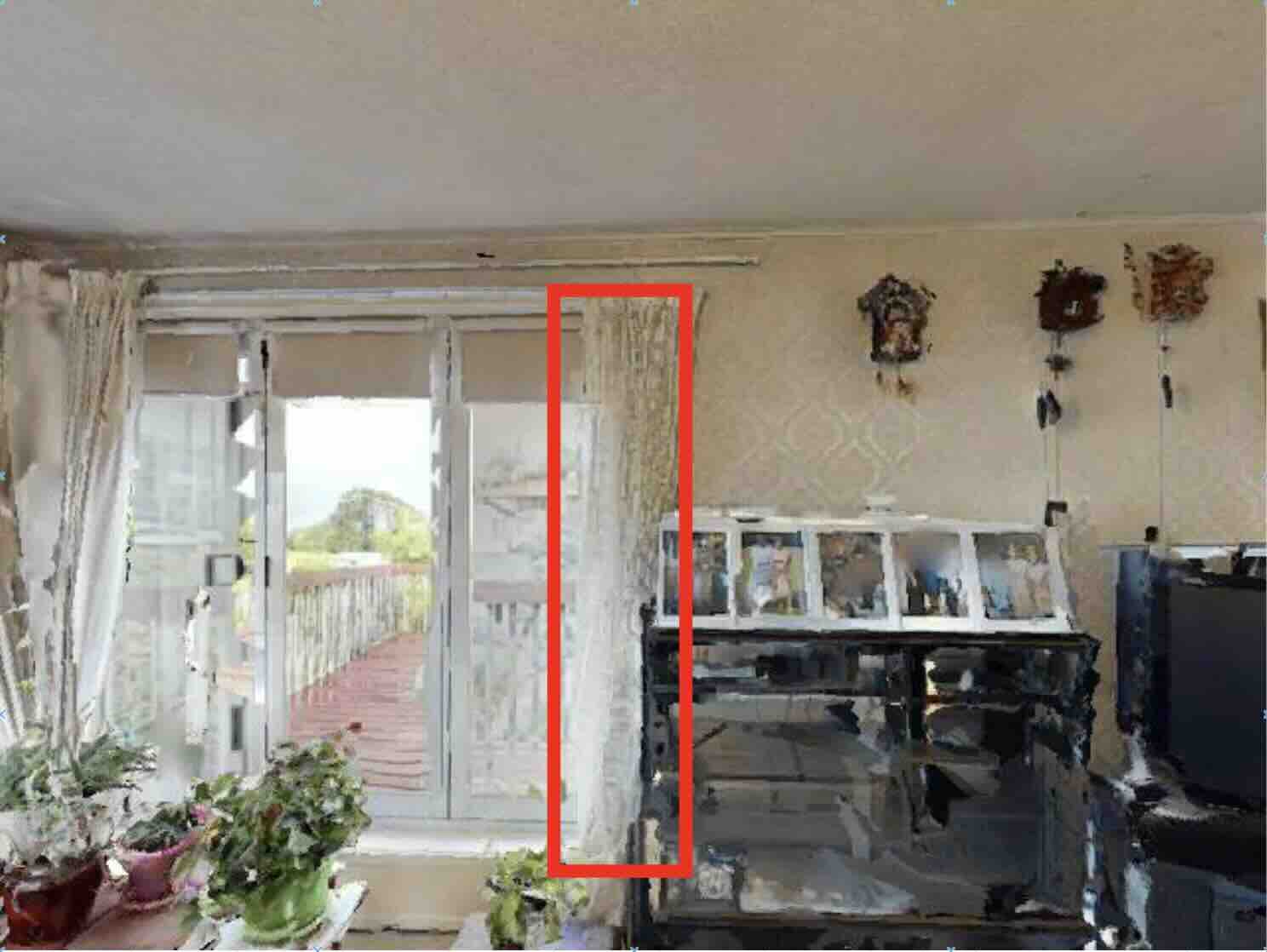} \\
                \textbf{(a)}
            \end{tabular} &
            \hspace{-8mm}
            \begin{tabular}{c}
                \includegraphics[width=0.5\linewidth]{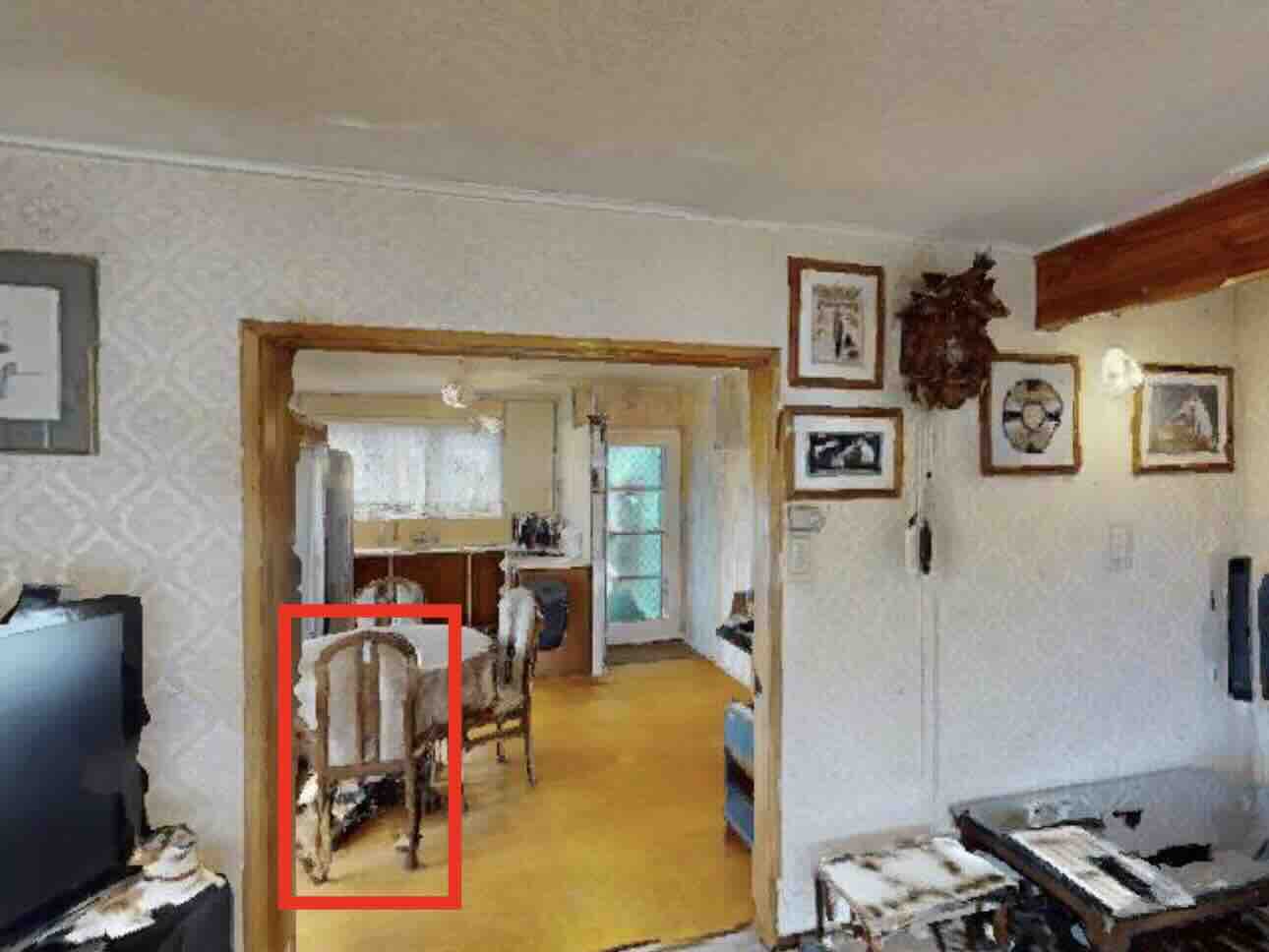} \\
                \textbf{(b)}
            \end{tabular}
        \end{tabular}
        \begin{tabular}{|p{0.95\linewidth}|}
            \hline
            \textbf{Ref:} ``Take off the right hand side white lace curtain and put it over the dining chair which is closest to the living room.'' \\
            \hline
            \textbf{Gemini:} ``Take the grandfather clock and place it on the kitchen counter.'' \\
            \\
            \hline
            \textbf{GPT-4o:} ``Take the plant in the blue pot from the windowsill and place it on the table in the dining area.'' \\
            \hline
            \textbf{Ours:} ``Move the white curtain hanging on the window and put it into the wooden chair in the kitchen.'' \\
            \hline
        \end{tabular}
        \vspace{-1mm}
        \begin{tabular}{p{0.95\linewidth}}
        \vspace{-3mm}
        \begin{center}
        $\mathrm{(iii)}$     
        \end{center}
        \end{tabular}
    \end{minipage}
    \vspace{-5mm}
    \caption{Successed samples with the proposed method. (a) and (b) show the target object image and the receptacle image, respectively. The bounding box indicates the target object and receptacle that are included in the references.}
    \label{fig:success_qualitative}
    \vspace{-6.5mm}
\end{figure*}

\vspace{-1.5mm}
\section{
    Experimental Results
    \label{experimental Results}
}
\vspace{-1.5mm}

\subsection{Quantitative Results
\label{experiments-quantitative}
}

\vspace{-1mm}

Table \ref{tab:result} shows the quantitative results of the baseline and proposed methods on the HM3D-FC dataset. 
The values in the table show the average and standard deviation over five trials.
We used Dense Caption Baseline, GRIT, BLIP-2, Gemini\cite{reid2024gemini}, and GPT-4o\cite{achiam2023gpt} as the baseline methods.
Dense Caption Baseline was the method used to generate dense captions for target object and receptacle images with BLIP-2 and fused them into instructions using an LLM.
We employed GRIT and BLIP-2 because they are representative methods for image captioning in the COCO\cite{lin2014microsoft} benchmark.
Moreover, we also used Gemini and GPT-4o because they are representative MLLMs that have been reported to produce favorable results in many vision \& language tasks.
We trained GRIT and BLIP-2 on HM3D-FC dataset. 
For models that can only handle a single input image, only the target object image was used as input. 
Experiments with the MLLM models, Gemini and GPT-4o, were conducted in a few-shot setting.

In this study, we used Polos, PAC-S\cite{sarto2023positive}, RefPAC-S\cite{sarto2023positive}, SPICE\cite{anderson2016spice}, CIDEr, and BLEU4 as the evaluation metrics.
Polos is a supervised automatic evaluation metric for image captioning models. Polos computes scores from multimodal inputs, using a parallel feature extraction mechanism that leverages embeddings trained through large-scale contrastive learning.
PAC-S and Ref-PAC-S evaluate captions in an unsupervised manner by computing their similarity with embeddings derived from fine-tuned CLIP. Ref-PAC-S uses reference-based metrics in its evaluation.
By contrast, PAC-S is a reference-free metric that calculates scores based solely on the similarities between image features and text features.
We used Polos and RefPAC-S because scores in these metrics correlate strongly with human judgment.
To evaluate the sentences generated by MLLMs appropriately, we also employed PAC-S. This is a representative reference-free evaluation metric that considers only the similarities between the generated sentences and images.
These evaluation metrics only calculate the similarity between an instruction and a single image. 
They cannot handle multiple images as simultaneous inputs.
Therefore, we independently calculated scores for the target object images and the receptacle images.
In addition, we also used SPICE, CIDEr, and BLEU4 because these are standard metrics in image captioning.

These results indicate that the proposed method outperformed BLIP-2 (which had the highest Polos scores of the baseline methods) by 8.5 points for the target object images and 10.6 points for the receptacle images.
Similarly, 
the proposed method outperformed the baseline methods with respect to PAC-S score for target object images and receptacle images, with improvements of 2.0, and 1.4 points compared with the best-performing baseline methods for each metric, respectively. 
It indicates that even with a representative reference-free evaluation metric, the proposed method outperformed all baseline methods.
Moreover, the proposed method outperformed in RefPAC-S, SPICE, CIDEr, and BLEU4 compared with the highest baseline in each metric.
There were statistically significant differences in the performances of our method and the baseline methods in terms of all evaluation metrics ($p$-value < 0.05).
\begin{figure}[t]
    \centering
    \begin{tabular}{c c}
        \begin{tabular}{c}
            \includegraphics[width=0.43\linewidth]{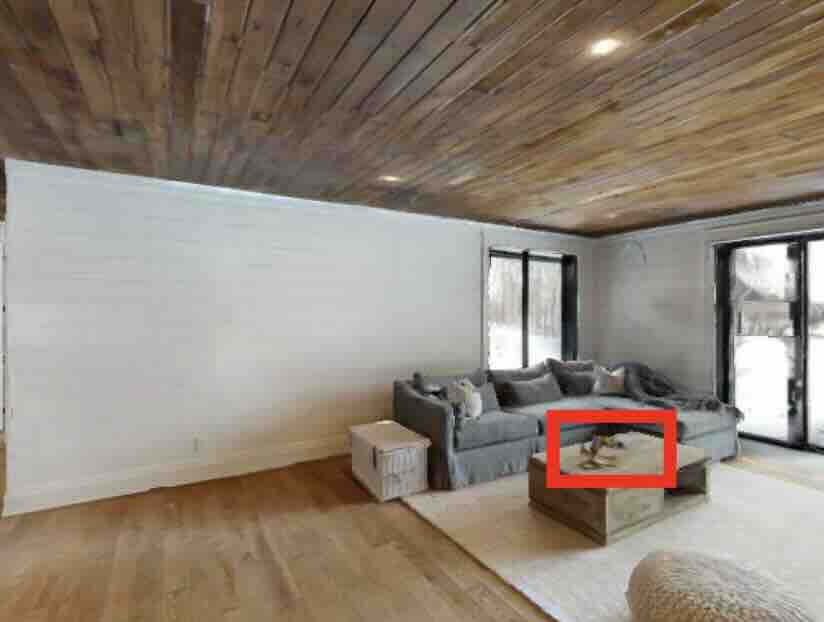} \\
            \vspace{-0.5mm}
            \textbf{(a)}
        \end{tabular} &
        \hspace{-0.6cm}
        \begin{tabular}{c}
            \includegraphics[width=0.43\linewidth]{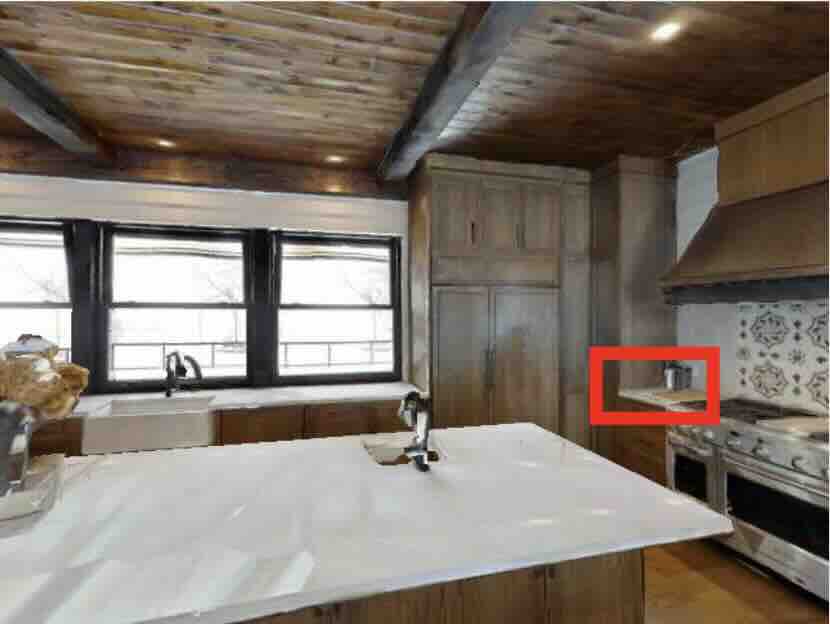} \\
            \vspace{-0.5mm}
            \textbf{(b)}
        \end{tabular} \\
    \end{tabular}
    
    \begin{tabular}{|p{0.9\linewidth}|}
        \hline
        \textbf{Ref:} ``Please bring back the small object on the wooden table to the right side corner of the kitchen.'' \\
        \hline
        \textbf{Ours:} ``Move the white cushion on the sofa to the shelf above the kitchen.'' \\
        \hline
    \end{tabular}     
    \caption{A failed samples obtained by the proposed method. (a) Target object image and (b) receptacle image, respectively.}
    \label{fig:fail_qualitative}
    \vspace{-6mm}
\end{figure}
\subsection{Qualitative Results}
Fig. \ref{fig:success_qualitative} shows successful examples obtained by using the proposed method.
In Fig. \ref{fig:success_qualitative} (i), the target object and receptacle were a red object and cupboard, respectively.
The reference was ``Move the red object on the sofa to the cupboard at the corner.''
Our method generated the sentence ``Move the red object on the sofa to the shelf above the kitchen.''
Thus, our method described the target object and the receptacle using appropriate colors and referring expressions. 
By contrast, BLIP-2 and GPT-4o output ``Move the white curtain on the left side of the window to the white shelf on the right side of the window.'' and ``Take the two glass figurines from the wooden ledge and place them on the kitchen counter next to the sink.'', respectively.
In these generated sentences, the descriptions of both the target object and the receptacle object were inappropriate.
In Fig. \ref{fig:success_qualitative} (ii), the target object and receptacle were a cushion and a chair, respectively.
The reference sentence for this example was ``Pick up the cushion on the sofa and put it on the chair near the window.'' 
The proposed method generated the sentence ``Could you move the brown cushion on the sofa to the wooden chair in the corner of the room?'', appropriately included both the target object and the receptacle in each image.
By contrast, BLIP-2 and GPT-4o output ``Could you move the cushion on the sofa and put it on the brown sofa?'' and ``Take the brown cushion from the couch and place it on the counter next to the red bottle in the kitchen.'', respectively.
In the sentence generated by BLIP-2, the descriptions of both the target object and the receptacle object represented the wrong color.
On the other hand, the generated sentence by GPT-4o refers to a receptacle with the nonexistence object `red bottle'.
In the case of Fig. \ref{fig:success_qualitative} (iii), the target object and receptacle were a curtain and a chair, respectively.
The reference was ``Take off the right hand side white lace curtain and put it over the dining chair which is closest to the living room.''
Our method generated the sentence ``Move the white curtain hanging on the window and put it into the wooden chair in the kitchen.'', describing the positions and materials of the target object and the receptacle appropriately. 
By contrast, Gemini and GPT-4o generated the sentences ``Take the grandfather clock and place it on the kitchen counter.’’ and ``Take the plant in the blue pot from the windowsill and place it on the table in the dining area.’’, respectively.
GPT-4o incorrectly recognized the color of the pot, while Gemini misidentified the small clock-like objects as a `grand-father clock'.

Fig. \ref{fig:fail_qualitative} shows a failed example of the proposed method.
In Fig. \ref{fig:fail_qualitative}, the target object and receptacle were a small object and right side corner of the kitchen, respectively. For this example, the reference sentence was ``Please bring back the small object on the wooden table to the right side corner of the kitchen.''
The sentence generated by the proposed method was ``Move the white cushion on the sofa to the shelf above the kitchen.'' 
The proposed method incorrectly described the target object and the receptacle as different from those in the reference captions. Moreover, it incorrectly described the color of the target object (cushions) as white, despite the target object image showing gray cushions.

\begin{table*}[hbt!]
\centering
\small
\setlength{\tabcolsep}{3.5pt}
\vspace{1mm}
\caption{\small Quantitative results of the ablation studies. Here, TQ refers to Triplet Qformer.}
\vspace{-2mm}
\normalsize
\begin{tabular}{l|p{1.5cm} p{1.5cm} p{1.5cm} p{1.5cm} p{1.5cm} p{1.5cm} c c c} 
    \hline
    \centering
    \multirow{2}{*}{Model} & \multicolumn{2}{c}{Polos$\uparrow$} & \multicolumn{2}{c}{PAC-S$\uparrow$} & \multicolumn{2}{c}{RefPAC-S$\uparrow$} & \multirow{2}{*}{SPICE$\uparrow$} & \multirow{2}{*}{CIDEr$\uparrow$} & \multirow{2}{*}{BLEU4$\uparrow$} \\ 
     & \centering \small{Target} & \centering \small{Receptacle} & \centering \small{Target} & \centering \small{Receptacle} & \centering \small{Target} & \centering \small{Receptacle} &  & &  \\ 
    \hline
    (i) \hspace{0.5mm} w/o HCCP & \centering \small{44.7 {$\scriptscriptstyle \pm 0.1$}} & \centering \small{44.4 {$\scriptscriptstyle \pm 0.2$}} & \centering \small{72.4 {$\scriptscriptstyle \pm 0.2$}} & \centering \small{70.6 {$\scriptscriptstyle \pm 0.2$}} & \centering \small{77.8 {$\scriptscriptstyle \pm 0.1$}} & \centering \small{76.7 {$\scriptscriptstyle \pm 0.1$}} & \centering \small{20.9 {$\scriptscriptstyle \pm 0.5$}} & \centering \small{53.3 {$\scriptscriptstyle \pm 1.6$}} & \small{9.0 {$\scriptscriptstyle \pm 0.5$}} \\ 
    (ii)\hspace{0.7mm} w/o TQ & \centering \small{46.2 {$\scriptscriptstyle \pm 0.2$}} & \centering \small{44.7 {$\scriptscriptstyle \pm 0.2$}} & \centering \small{74.2 {$\scriptscriptstyle \pm 0.2$}} & \centering \small{69.9 {$\scriptscriptstyle \pm 0.2$}} & \centering \small{78.6 {$\scriptscriptstyle \pm 0.1$}} & \centering \small{76.0 {$\scriptscriptstyle \pm 0.1$}} & \centering \small{18.3 {$\scriptscriptstyle \pm 0.3$}} & \centering \small{55.3 {$\scriptscriptstyle \pm 1.4$}} & \small{10.2 {$\scriptscriptstyle \pm 0.2$}} \\ \hline
    (iii) full & \centering \small{\textbf{50.9}} {$\scriptscriptstyle \pm 0.8$} & \centering \centering \small{\textbf{{50.7}} {$\scriptscriptstyle \pm 0.7$}} & \centering \small{{\textbf{74.3}} {$\scriptscriptstyle \pm 0.6$}} & \centering \small{{\textbf{72.1}} {$\scriptscriptstyle \pm 0.7$}} & \centering \small{{\textbf{79.8}} {$\scriptscriptstyle \pm 0.4$}} & \centering \small{{\textbf{78.5}} {$\scriptscriptstyle \pm 0.5$}} & \centering \small{\textbf{22.7} {$\scriptscriptstyle \pm 0.4$}} & \centering \small{\textbf{64.8} {$\scriptscriptstyle \pm 4.0$}} & \small{\textbf{11.5} {$\scriptscriptstyle \pm 1.1$}} \\ \hline
\end{tabular}

\label{tab:ablation}
\vspace{-5.5mm}
\end{table*}

\vspace{-1mm}
\subsection{Ablation Studies}
Table \ref{tab:ablation} shows the results of the ablation studies. 
The following two ablation conditions were set:

\textbf{HCCP Ablation:} The proposed method was trained using the three stages. 
The effectiveness of the human-centric calibration phase (HCCP) was then investigated by removing this stage. 
As shown in Table \ref{tab:ablation}, Model (i) achieved Polos scores of 44.7 and 44.4 for the target object images and receptacle images, respectively.
Compared with Model (iii), the Polos scores obtained were 6.2 and 6.3 points lower.
In fact, Model (iii) outperformed Model (i) in all evaluation metrics.
These results indicate that training with the HCCP improved the quality of instruction sentences, allowing them to more closely resemble instructions given by humans.

\textbf{Triplet Qformer Ablation:} The effectiveness of Triplet Qformer was demonstrated by replacing it with a standard Qformer\cite{li2023blip2}. 
In this comparison, the concatenated features of $\bm{H}_\mathrm{G}$ and $\bm{H}_\mathrm{R}$ were used as the input to the Qformer in Model (ii).
According to Table \ref{tab:ablation}, Model (ii) achieved Polos scores of 46.2 and 44.7 for the target object images and receptacle images, respectively.
Compared with Model (iii), the Polos scores decreased by 4.7 and 6.0 points, respectively.
In general, Model (iii) outperformed Model (ii) across all evaluation metrics.
These results reveal that the structure of Triplet Qformer enables the two types of visual features to be appropriately aligned with text features.

\vspace{-1mm}
\subsection{Dataset Augmentation Experiments
\label{dataset-augumentation-experiments}
}
\vspace{-1mm}
\begin{table}[hbt]
    \vspace{0.8mm}
    \caption{\small Quantitative results of the data augmentation experiment. (*-T) and (*-R) denote the models for the target objects and the receptacles, respectively. The numbers in bold indicate the best scores for each image in each metric.}
    \normalsize
    \vspace{-1mm}
    \label{tab:quant_dataset}
        \begin{tabular}{@{\hspace{1.5mm}}l@{\hspace{1.5mm}}cc@{\hspace{1.5mm}}c@{\hspace{1.5mm}}c@{\hspace{1.5mm}}c@{\hspace{1.5mm}}c@{\hspace{1.5mm}}}
            \toprule
             \multicolumn{1}{c}{\multirow{2}{*}{[\%]}} & \multicolumn{1}{c}{\multirow{2}{*}{Condition}} & \multicolumn{4}{c}{HM3D-FC (unseen)} \\
             \cmidrule{3-6}
            & & MRR$\uparrow$ & R@5$\uparrow$ & R@10$\uparrow$ & R@20$\uparrow$ \\
            \midrule
            (i-T) & \multirow{2}{*}{half} & $14.9 \scriptscriptstyle{\pm 2.0}$ & $19.0 \scriptscriptstyle{\pm 3.7}$ & $37.1 \scriptscriptstyle{\pm 3.9}$ & $64.3 \scriptscriptstyle{\pm 4.9}$ \\
            (i-R) &  & $17.9 \scriptscriptstyle{\pm 1.1}$ & $25.3 \scriptscriptstyle{\pm 3.5}$ & $43.2 \scriptscriptstyle{\pm 3.4}$ & $70.3 \scriptscriptstyle{\pm 1.7}$ \\
            (ii-T) & \multirow{2}{*}{half + Aug} & $19.5 \scriptscriptstyle{\pm 2.4}$ & $27.0 \scriptscriptstyle{\pm 4.2}$ & $44.2 \scriptscriptstyle{\pm 3.7}$ & $69.8 \scriptscriptstyle{\pm 4.3}$ \\
            (ii-R) &  & $18.7 \scriptscriptstyle{\pm 2.0}$ & $26.8 \scriptscriptstyle{\pm 4.3}$ & $44.5 \scriptscriptstyle{\pm 4.4}$ & $70.1 \scriptscriptstyle{\pm 4.0}$ \\
            \midrule
            (iii-T) & \multirow{2}{*}{full} & $20.5 \scriptscriptstyle{\pm 2.3}$ & $30.1 \scriptscriptstyle{\pm 3.4}$ & $48.2 \scriptscriptstyle{\pm 1.4}$ & $73.2 \scriptscriptstyle{\pm 2.8}$ \\
            (iii-R) &  & $19.8 \scriptscriptstyle{\pm 1.1}$ & $27.1 \scriptscriptstyle{\pm 3.2}$ & $49.1 \scriptscriptstyle{\pm 5.9}$ & $74.6 \scriptscriptstyle{\pm 3.1}$ \\

            (iv-T) & \multirow{2}{*}{full + Aug} & $\mathbf{22.3} \scriptscriptstyle{\pm 1.8}$ & $\mathbf{31.9} \scriptscriptstyle{\pm 4.4}$ & $\mathbf{50.2} \scriptscriptstyle{\pm 3.5}$ & $\mathbf{75.1} \scriptscriptstyle{\pm 4.3}$ \\
            (iv-R) & & $\mathbf{22.1} \scriptscriptstyle{\pm 1.2}$ & $\mathbf{32.0} \scriptscriptstyle{\pm 3.8}$ & $\mathbf{52.3} \scriptscriptstyle{\pm 1.9}$ & $\mathbf{77.3} \scriptscriptstyle{\pm 3.4}$ \\
            \bottomrule
        \end{tabular}
    \vspace{-4mm}
\end{table}

To investigate the effectiveness of the instruction sentences generated by the proposed method for data augmentation, we conducted data augmentation experiments.
In these experiments, we focused on the Image Retrieval-based Open-Vocabulary Fetch-and-Carry (IROV-FC) task \cite{korekata24arxiv}.
For this task, a system in which a robot identifies and retrieves images of the specified target object and receptacle based on a free-form mobile manipulation instruction must be developed. 
This task was selected because of the applicability of the sentences generated by the proposed method.
We used mean reciprocal rank (MRR) and recall@$K$ (R@$K$) because they are standard metrics used in machine learning problems that involve ranking\cite{kaneda2024learning, korekata24arxiv}.
We used MultiRankIt\cite{kaneda2024learning} as an IROV-FC model because it is a representative IROV-FC model. We trained this model on the LTRRIE-FC dataset\cite{korekata24arxiv} which is a superset of the HM3D-FC dataset.
Here, we trained MultiRankIt for target objects and receptacles separately as it cannot handle both modes within a single model. Therefore, we evaluated MultiRankIt for target objects and receptacles individually, and then, calculated the average.

We created the Aug-train dataset with instruction sentences generated using images in the training set of the HM3D-FC dataset.
To build the Aug-train dataset, we attached a single instruction sentence generated by the proposed method to each image included in the training set of the HM3D-FC dataset. Here, the proposed method was trained on HM3D-FC.
We set the following four conditions in the data augmentation experiment:
\begin{itemize}
    \item[(i)] \textbf{half}: Half of the training set of the LTRRIE-FC dataset.
    \item[(ii)] \textbf{half + Aug}: Half of the training set of the LTRRIE-FC dataset (randomly chosen) plus the Aug-train dataset.
    \item[(iii)] \textbf{full}: All of the training samples in the training set of the LTRRIE-FC dataset.
    \item[(iv)] \textbf{full + Aug}: The full  training set of the LTRRIE-FC and Aug-train datasets.
\end{itemize}

Table \ref{tab:quant_dataset} shows the quantitative results.
From the table,
Models (iv-T) and (iv-R) achieved MRRs of 22.3\% and 22.1\%, whereas Models (iii-T) and (iii-R) achieved 20.5\% and 19.8\%, respectively.
These results indicate that Models (iv-T) and (iv-R) outperformed Models (iii-T) and (iii-R) by 1.8 points and 2.3 points in terms of MRR on the HM3D-FC test set.
Similarly, the models with augmented datasets outperformed those trained without augmented datasets in terms of R@$K$.
We confirmed that performing data augmentation on language instructions with our model improved the performance of the existing IROV-FC model.

\vspace{-2mm}
\subsection{Physical Experiments}
\vspace{-1mm}
In this experiment, we confirmed that dataset augmentation using the proposed method is effective even in the physical experiments.
The experimental environment and robot platform were set up according to \cite{korekata24arxiv}.
We validated the effects of data augmentation using MultiRankIt trained on conditions (iii) and (iv), as described in Section \ref{dataset-augumentation-experiments}.
In the experiments, the robot executed mobile manipulation based on free-form instructions given by the users.
The experiments were conducted over 20 episodes: four different environmental settings times five episodes in per setting.
We performed the remaining settings according to the physical experiment in \cite{korekata24arxiv}.
\begin{table}[t]
    \caption{\small Quantitative results for the physical experiments.}
    \vspace{-1mm}
    \label{tab:quant_physical}
    \normalsize
    \centering
    \resizebox{0.9\columnwidth}{!}{
        \begin{tabular}{ccccc}
            \toprule
            $[\%]$ & Condition & MRR$\uparrow$ & R@10$\uparrow$ & SR$\uparrow$ \\
            \midrule
            (iii) & full & 21 & 51 & 45 (9/20) \\
            (iv) & full + Aug & \textbf{23} & \textbf{58} & \textbf{55} (11/20) \\
            \bottomrule
        \end{tabular}
    }
    \vspace{-5mm}
\end{table}

We used MRR, $\text{R@}10$ and task success rate (SR) as the evaluation metrics for the physical experiments.
Table \ref{tab:quant_physical} shows that Model (iv) achieved SR of 55\%, which was 10 points higher than the unaugmented model.
These results indicate that data augmentation on language instructions with our model contributed to improving the performance of MultiRankIt, leading to an increase in the SR of fetching and carrying actions.

\vspace{-1.5mm}
\vspace{-1mm}
\section{Conclusions}
\vspace{-1.5mm}

In this study, we focused on the mobile manipulation instruction generation task.
The contributions of this study were as follows.
We introduced Triplet Qformer, which enabled the alignment of multiple types of visual features individually with text features and then used the text features as an anchor to align the visual features with each other. We also introduced human centric calibration phase, a training method that uses human centric calibration training loss function with learning-based and $n$-gram-based automatic evaluation metrics. Our proposed method outperformed the baseline methods including representative MLLMs on automatic evaluation metrics on the HM3D-FC dataset. We confirmed that performing data augmentation on language instructions with our model improved the performance of an existing Image Retrieval-based Open-Vocabulary Fetch-and-Carry model in both the dataset augmentation and physical experiments.

There were still some samples with color-related errors in the generated sentences by the proposed method.
 Therefore, introducing color detectors that label the color of each detected object and use it as an additional feature is one task for future work.


\vspace{-0.9mm}

\section*{ACKNOWLEDGMENT}
\vspace{-1.2mm}

This work was partially supported by JSPS KAKENHI Grant Number 23K28168, JST Moonshot, NEDO, and JSPS Fellows Grant Number JP23KJ1917.

\vspace{-1.5mm}
\bibliographystyle{IEEEtran}
\bibliography{reference}

\end{document}